\documentclass[conference]{IEEEtran}
\usepackage{times}

\usepackage[numbers]{natbib}
\usepackage{multicol}
\usepackage{multirow}
\usepackage{hyperref}
\usepackage{amsmath}
\usepackage{amssymb}
\usepackage{xcolor}
\usepackage{optidef}
\usepackage{booktabs}
\usepackage{bm}
\usepackage{enumitem}
\usepackage{todonotes}
\usepackage{lipsum}
\usepackage{capt-of,etoolbox}
\usepackage{mwe}
\usepackage{svg}

\renewcommand{\bold}[1]{\mathbf{#1}} 
\newcommand{\R}[0]{\mathbb{R}} 

\renewcommand{\vec}[1]{\mathbf{#1}}

\newcommand{\mat}[1]{\mathbf{#1}}

\newcommand{\mass}[0]{m}

\newcommand{\massGrasped}[0]{\mass_\text{go}}
\newcommand{\massExtrinsic}[0]{\mass_\text{eo}}

\newcommand{\stateSub}[1]{\bold x_\text{#1}}
\newcommand{\stateValue}[1]{\begin{bmatrix} x_\text{#1} & y_\text{#1} & \theta_\text{#1} \end{bmatrix}^\top}

\newcommand{\robotState}[0]{\stateSub{ee}}
\newcommand{\robotStateSub}[1]{\stateSub{ee#1}}
\newcommand{\robotStateValue}[0]{\stateValue{ee}}
\newcommand{\graspedObjectState}[0]{\stateSub{go}}
\newcommand{\graspedObjectStateSub}[1]{\stateSub{go#1}}
\newcommand{\graspedObjectStateValue}[0]{\stateValue{go}}
\newcommand{\extrinsicObjectState}[0]{\stateSub{eo}}
\newcommand{\extrinsicObjectStateSub}[1]{\stateSub{eo#1}}
\newcommand{\extrinsicObjectStateValue}[0]{\stateValue{eo}}

\newcommand{\contactPoint}[1]{\bold r_\text{c,#1}}
\newcommand{\graspedContactPoint}[0]{\contactPoint{obj}}
\newcommand{\graspedContactPointSub}[1]{\contactPoint{obj, #1}}

\newcommand{\externalWrench}[0]{\bold w_\text{ext}}

\newcommand{\seq}[1]{\{#1_{k}\}_{k=1}^K}

\newcommand{\contactMode}[0]{c}
\newcommand{\contactModeSeq}[0]{\seq{\contactMode}}

\begin{document}

\title{Tactile-Driven Non-Prehensile Object Manipulation via Extrinsic Contact Mode Control}


\author{
Miquel Oller\hspace{15pt} Dmitry Berenson \hspace{15pt} Nima Fazeli\\
Department of Robotics, University of Michigan \\
Ann Arbor, MI 48109, United States\\
\texttt{\{oller, dmitryb, nfz\}@umich.edu} \\
\url{https://www.mmintlab.com/tactile-nonprehensile} 
}



%

\makeatletter
\let\@oldmaketitle\@maketitle
\renewcommand{\@maketitle}{\@oldmaketitle
\bigskip
    \begin{center}
  \includegraphics[width=0.8\linewidth]{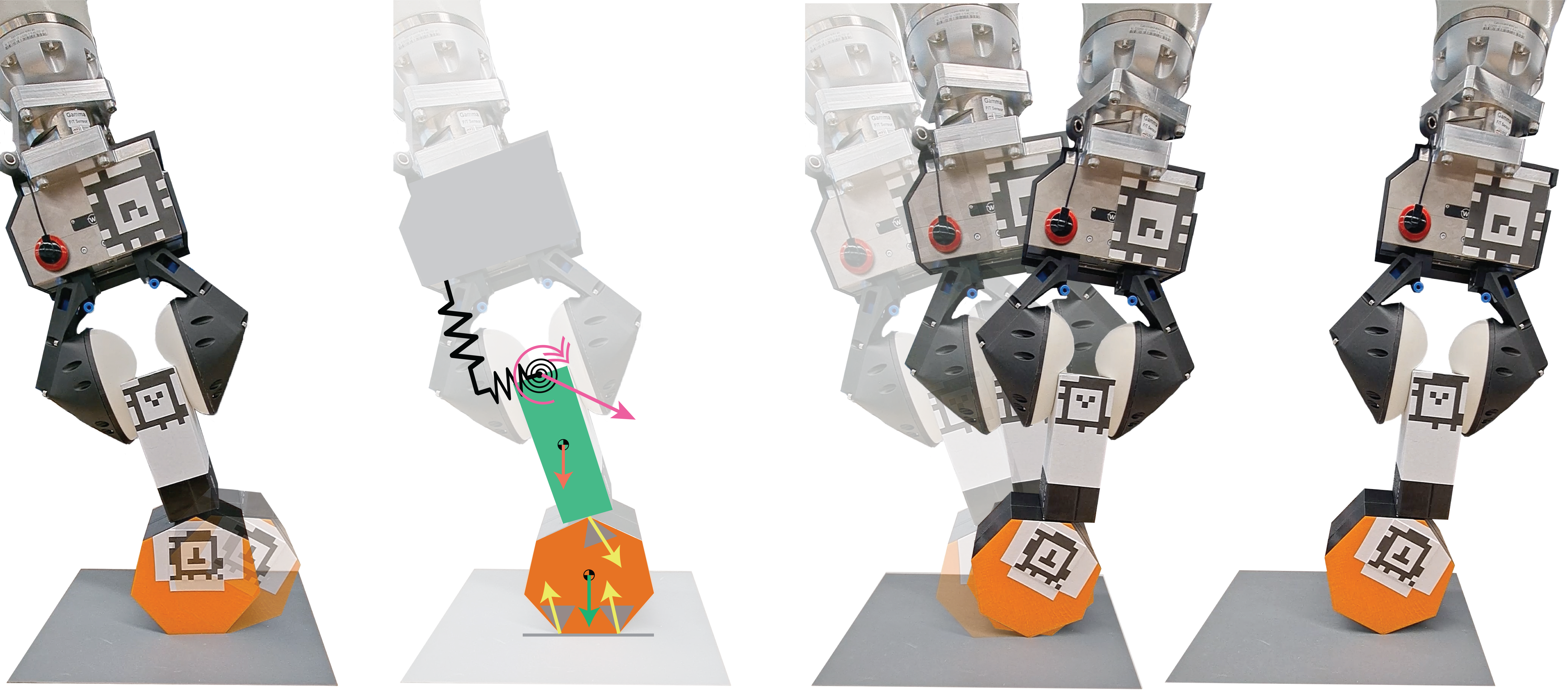}
    \setcounter{figure}{0}
    \captionof{figure}{\textbf{Overview of Approach:} \textit{First Panel:} our general setup where the robot is grasping an object with tactile sensors and is tasked with manipulating an object in the environment. In this example, the goal of the robot is to pivot the object resting on the surface. \textit{Second Panel:} a schematic of our robot's understanding of the world. The robot models the tactile sensors using passive compliance elements, and considers the frictional interaction between the objects in the scene. \textit{Third Panel:} the executed trajectory from our planner. \textit{Fourth Panel:} the robot successfully drives the objects to their goal poses.}
    \label{fig:teaser}
    \end{center}
    }
\makeatother

\maketitle

\begin{abstract}

In this paper, we consider the problem of non-prehensile manipulation using grasped objects. This problem is a superset of many common manipulation skills including instances of tool-use (e.g., grasped spatula flipping a burger) and assembly (e.g., screwdriver tightening a screw). Here, we present an algorithmic approach for non-prehensile manipulation leveraging a gripper with highly compliant and high-resolution tactile sensors. Our approach solves for robot actions that drive object poses and forces to desired values while obeying the complex dynamics induced by the sensors as well as the constraints imposed by static equilibrium, object kinematics, and frictional contact. Our method is able to produce a variety of ``manipulation skills'' and is amenable to gradient-based optimization by exploiting differentiability within contact modes (e.g., specifications of ``sticking'' or ``sliding'' contacts). We evaluate 4 variants of controllers that attempt to realize these plans and demonstrate a number of complex skills including non-prehensile planar sliding and pivoting on a variety of object geometries. The perception and controls capabilities that drive these skills are the building blocks towards dexterous and reactive autonomy in unstructured environments. 
%
\end{abstract}

\IEEEpeerreviewmaketitle

\section{Introduction}





Precise control over contact interactions is an essential capability for dexterous manipulation \cite{mason_manipulation}. Consider, for instance, the intricate task of pivoting or balancing an object on the table using a grasped object (tool), Fig.~\ref{fig:teaser}. To execute this task effectively, the robot must reason about the precise physical interactions between the various elements involved: the connection between its end-effector and the tool, the interaction between the tool and the object, and the interplay between the object and the table. If done correctly, the robot is imbued with a myriad of dexterous capabilities. However, there are also many ways in which the robot can fail, including failing to generate sufficient contact force through friction to lift the object, the object sticking to the table due to friction or sliding away from the tool, or the tool slipping out of the robot's grasp. These failures are due to the nonlinear, discontinuous, and multimodal nature of contact interactions.



In this paper, we consider the class of problems in which the robot is tasked with using an object grasped with tactile sensors to: i) transmit desired forces to the environment, and ii) manipulate other extrinsic objects: objects for which the robot does not have a direct grasp. Fig.~\ref{fig:teaser} shows an example of the system where the robot is grasping the box using tactile sensors and is to pivot the hexagon.
Our contribution is a gradient-based trajectory optimization algorithm for extrinsic object manipulation that exploits the passive compliance of the tactile sensors for quasi-static in-contact force transmission. More specifically, our approach solves for robot actions that drive object poses and forces to desired values while obeying the complex dynamics induced by the sensors as well as the constraints imposed by static equilibrium, object kinematics, and frictional contact. Our method is able to produce a variety of ``manipulation skills'' and is amenable to gradient-based optimization by exploiting differentiability within contact modes (e.g., specifications of ``sticking'' or ``sliding'' contacts). We demonstrate our approach on a number of dexterous extrinsic contact and object manipulation skills including grasped object pivoting as well as extrinsic object pushing and pivoting.

An integral part of our method is the use of tactile sensors. This is because robots of the future will likely extensively use tactile sensors as they provide a means for feedback directly from the physical interaction between the robot and its environment. This feedback can be used for both perception (as we do here for object localization and extrinsic contact detection) as well as for controls. Here, we use high-resolution and highly deformable tactile sensors (Soft Bubbles \cite{soft_bubbles}) because they: i) allow for state-estimation that provides key feedback for controls that would not be available without the sensors, and ii) their compliance facilitates contact-rich interaction without the need for extensive hand-tuning and precise impedance control from the robot arm. This makes our approach more robust to uncertainty and accessible given the lower technical barrier to entery. With small modifications detailed in Sec.~\ref{sec:methods}, our approach can also apply to more rigid tactile sensors (e.g., GelSight \cite{yuan2017gelsight} GelSlims \cite{gelslim3}, Fingervision \cite{yamaguchi2017implementing}, DIGIT \cite{lambeta2020digit}) together with impedance control.

\section{Related Work}

\subsection{Tactile State Estimation}
Tactile sensing plays a crucial role in contact-rich manipulation tasks. It provides direct feedback about the forces, pressures, and contact points between the robot and the objects it interacts with \cite{alberto_unstable_queen}. This feedback can solely be used to determine the object pose \cite{bauza_tac2pose, tactile_pose_policy, bubble_pose_estimation} or can complement other sensing modalities, such as vision \cite{visuotactile_6d_pose, vision_tactile_servoing}.

Analogous to the effectiveness of wrench feedback in resolving contact locations \cite{contact_particle_filter, multiscope},
advanced high-resolution tactile sensors like Gelslim \cite{gelslim3} or Soft Bubbles \cite{soft_bubbles} offer comparable sensing capabilities. \citet{daolin_contact_sensing} demonstrated this by integrating high-resolution tactile feedback with rigid-body and non-penetration constraints to estimate contact locations on a grasped object.
Expanding on this approach, \citet{kim2022active} introduced factor graphs for effectively combining contact constraints and sensor measurements. Furthermore, 
\citet{neural_contact_fields} innovatively combined tactile feedback with implicit geometry representations and proprioception to estimate the extrinsic contact across a diverse set of objects and contact geometries.

In this work, we leverage tactile sensors to estimate both the in-hand object pose of the grasped object and the contact location between the grasped object and external entities.

\subsection{Forceful Tactile Control}

Controlling the transmitted forces during manipulation tasks is key for tool use and other manipulation tasks like opening bottles \cite{rachel_forceful}.
Tactile sensors have been shown useful to control such forceful interaction between the robot and the grasped object. \citet{kim2023simultaneous} use factor graph formulation to estimate and control extrinsic contacts with tactile feedback.
\citet{manipulation_via_membranes} learn the sensor deformation dynamics for manipulating grasped objects to attain a target pose and transmitted wrenches to the environment.
Other models have been proposed to model the Soft Bubbles deformation-force relationship. \citet{bubble_hydroelastic_model} proposed to consider the sensors as a hydro-elastic element to link deformation and pressure.
Similarly, \citet{seed} proposes to model the object-sensor interaction as a Series of Elastic End-effectors (SEED). This model only requires estimating a few parameters and turns a force control problem into a more manageable position control, suitable for industrial robots.
In this method, we follow this later approach to model the sensor-object interaction. Since our study case is planar motions, we simplify the SEED model into two linear and one rotational spring.

\subsection{Contact-rich Manipulation Planning and Control}

Explicitly reasoning about the contact interaction is key for devising complex locomotion and manipulation strategies \cite{posa2014direct, manchester2020variational, mordatch2012contact, bernardo_quasidynamic_contact}. However, the discrete and multi-modal nature of contact poses a significant challenge in generating contact-rich trajectories. \citet{nikhil_planar_motion_cones} proposed a planning framework that considers a generalization of motion cones to generate in-contact feasible motions for prehensile manipulation. 
\citet{fast_pivoting_planning} and 
\citet{hogan_tactile_dexterity} dissected a manipulation task into sequences of primitive motions. Similarly, \citet{mason_hierarchical_contact} presents a hierarchical planning framework for planning through rigid body motions and complex contact sequences based on Monte Carlo tree search.

Controlling the extrinsic contact modes can serve to in-hand readjustment of grasped objects \cite{nikhil_extrinsic_dexterity,  nikhil_stable_prehensile_pushing}, or manipulating objects when a direct grasp is not possible \cite{manipulation_shared_grasping, robust_pivoting_shirai}.
Building upon this domain, 
\citet{doshi2022manipulation} considered a manipulation of an object in contact with the robot on a plane. They devised a contact configuration estimator and mode-dependent controller to manipulate unknown objects.
\citet{taylor2023object} extended this formulation by introducing factor graphs for estimating the contact configuration constraints. 
Our method extends these methods to consider extrinsic object manipulation. In our case, the robot uses a grasped object to manipulate another planar object.

Closely related to our work, \citet{tactile_tool_manipulation} presents a trajectory optimization scheme for pivoting an extrinsic object using a grasped tool. The main difference between this approach and ours is that we consider different primitive motions beyond pivoting, like sliding and pushing. Additionally, the rigid contact assumption on the robot-object does not hold for the Soft Bubbles sensors due to the substantial magnitude of their deformations. Our approach addresses this challenge by incorporating a model that accounts for the deformation-force relationship inherent in these sensors. Further, our approach is also able to reason over force transmissions as observed variables due to the sensor compliance models.

\section{Problem Statement}
\begin{figure}[t!]
    \begin{minipage}[c]{.2\textwidth}%
    \centering
    \includegraphics[width=\textwidth]{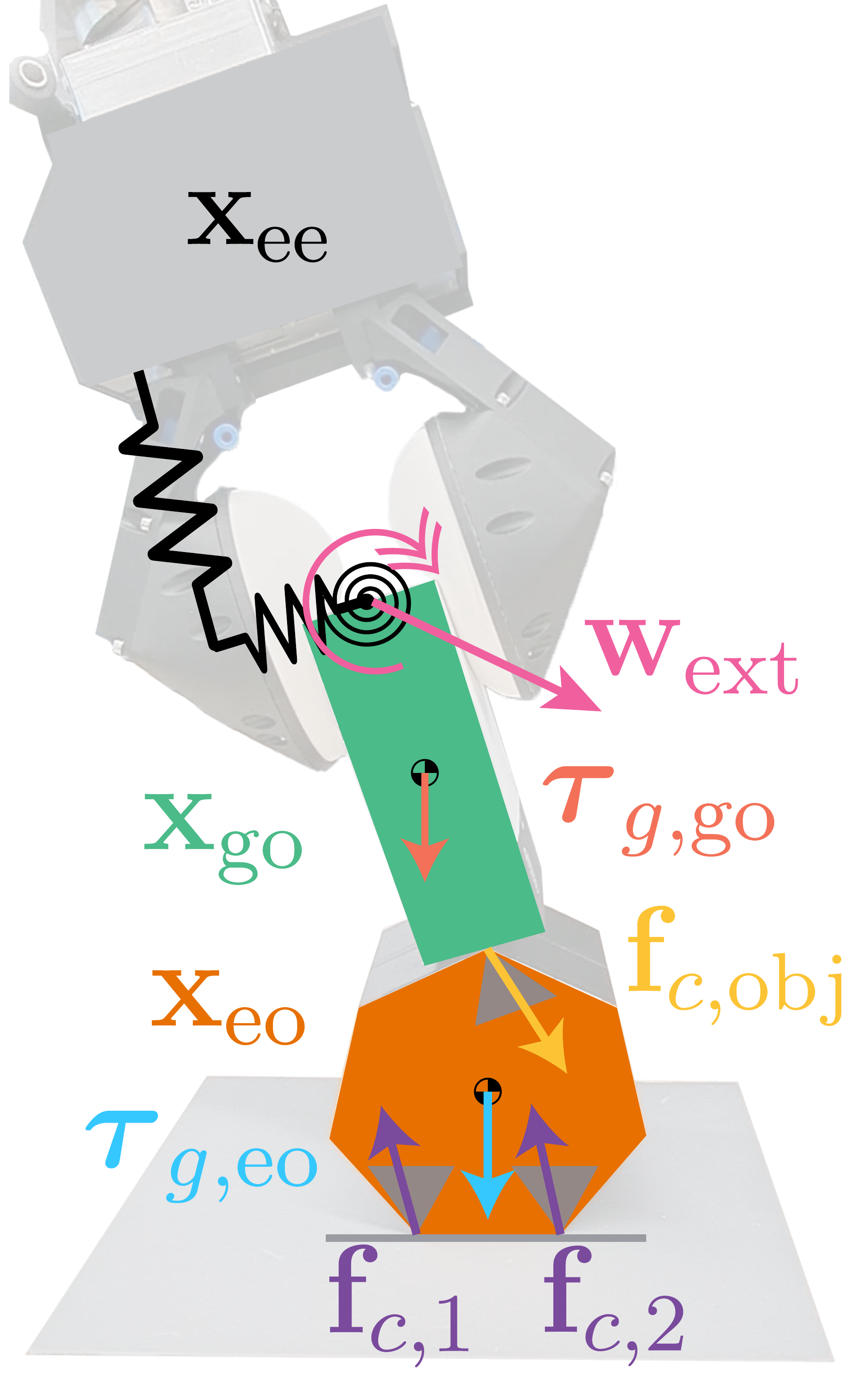}
    
    \end{minipage}
    \begin{minipage}[c]{.2\textwidth}
    \begin{center}
    \begin{tabular}{cc}
    \toprule
    Symbol & Description \\ 
    \cmidrule(lr){1-1} \cmidrule(lr){2-2}
      $\mathbf{x}_\text{ee}$ & End-effector Pose \\
      $\mathbf{x}_\text{go}$ & Grasped Object Pose \\
      $\mathbf{x}_\text{eo}$ & Extrinsic Object Pose \\
      $\mathbf{w}_\text{ext}$ & External Wrench \\
      $\bold{f}_{c,i}$ & Contact Force \\ 
      $\boldsymbol{\tau}_{g,\text{go}}$ & GO Gravitational Force \\
      $\boldsymbol{\tau}_{g,\text{eo}}$ & EO Gravitational. Force \\
      $m_\text{go}$ & Grasped Object Mass \\
      $m_\text{eo}$ & Extrinsic Object Mass \\
      $\mu_{i}$ & Contact Friction \\
      $c_{i}$ & Contact Mode \\
      $\graspedContactPoint$ & GO Contact Point \\
    \bottomrule
    \end{tabular}
     
    \end{center}
    \end{minipage}
    \label{fig:notation}
    \caption{\textbf{Notation}}
\end{figure}

We consider quasi-static manipulation of extrinsic polygonal planar objects of known geometry using a grasped object. The system is composed by 4 main elements:

\begin{itemize}
    \item The robot hand, equipped with Soft Bubbles tactile sensors \cite{soft_bubbles}, as seen in Fig.~\ref{fig:teaser}. These sensors are composed of an inflated membrane that is perceived from the inside of the finger via a time-of-flight depth camera. The sensor operation principle is to convert membrane deformations to high-resolution depth images.
    \item The grasped object, which is treated as a planar rigid polygon of mass $\massGrasped$. The grasped object moves with respect to the gripper because of the sensor membrane compliance. However, we assume that the contact between the object's surface and the membrane is sticking -- i.e., there is no slip between the surface of the object in contact with the membrane and the membrane. 
    \item The extrinsic object, which is treated as a planar rigid polygon of mass $\massExtrinsic$ moves in the plane of gravity.
    \item The ground plane, which is a fixed horizontal line.
\end{itemize}

The system states consist of the robot end-effector planar pose $\robotState=\robotStateValue$, the grasped object pose $\graspedObjectState=\graspedObjectStateValue$, and the extrinsic object pose $\extrinsicObjectState=\extrinsicObjectStateValue$. 

Given a feasible desired trajectory of the extrinsic object poses $\{\extrinsicObjectStateSub{,k}\}_{k=1}^K$ as well as the contact modes $\contactModeSeq$ along this sequence, our goal is to find a sequence of robot end-effector poses $\{\robotStateSub{,k}\}_{k=1}^K$ that minimizes error between both the desired object poses and forces transmitted with respect to their measured values. The contact modes define whether each contact ``sticks'', i.e. contact points between two objects move together and are fixed, or ``slips'', i.e. there is relative motion between the contact points. The key technical challenges are computing trajectories that obey the many unilateral and hybrid contact constraints, kinematic constraints imposed by geometry, accounting for the compliance of the sensors and maintaining stable object configurations.

\textbf {Assumptions:} In our approach, we make the following assumptions: First, we assume planar quasi-static rigid-body dynamics of known polygonal objects governed by Coulomb friction. We further assume that grasped object does not slide with respect to the tactile sensors (though it is able to move with respect to the end-effector frame due to compliance of the sensors). Finally, we assume that the grasped object interacts with the extrinsic object through as single point contact.

\section{Methodology}\label{sec:methods}

Our method is composed of 4 core components: i) a state-estimation pipeline using the feedback from the tactile sensor to estimate object pose and extrinsic contacts; ii) a passive compliance model for the tactile sensor that maps changes in grasped object pose to forces transmitted to the gripper; iii) the model used to predict the motion of the extrinsic object, and iv) a trajectory optimization approach to plan through the contacts necessary to achieve desired object poses and force transmissions. The main contributions of our work are in components (iii) and (iv) where we augment the model in (ii) with contact-aware constraints for object poses and force transmission, then formulating our planner and manipulation skills. Next, we provide details of these components:

\subsection{Tactile State-Estimation}
\label{sec:tactile_state_estimation}

In this section, we detail our algorithm for estimating the current grasped object pose $\graspedObjectState$ and contact location $\graspedContactPoint$ from tactile measurements. Starting from our assumption that the grasped object is a polygon of known geometry, we implement an edge detection algorithm that operates on the tactile sensor deformation field. Given the known object geometry and perceived edges in contact with the sensor, we fit the object pose following the alignment step from Iterative Closest Points (ICP) in which we minimize the L2 distance between the corresponding edge points. We constraint the object pose to be in $\text{SE}(2)$, i.e. in the task plane.

\begin{figure}[t!]
    \centering
    \includegraphics[width=0.5\textwidth]
    {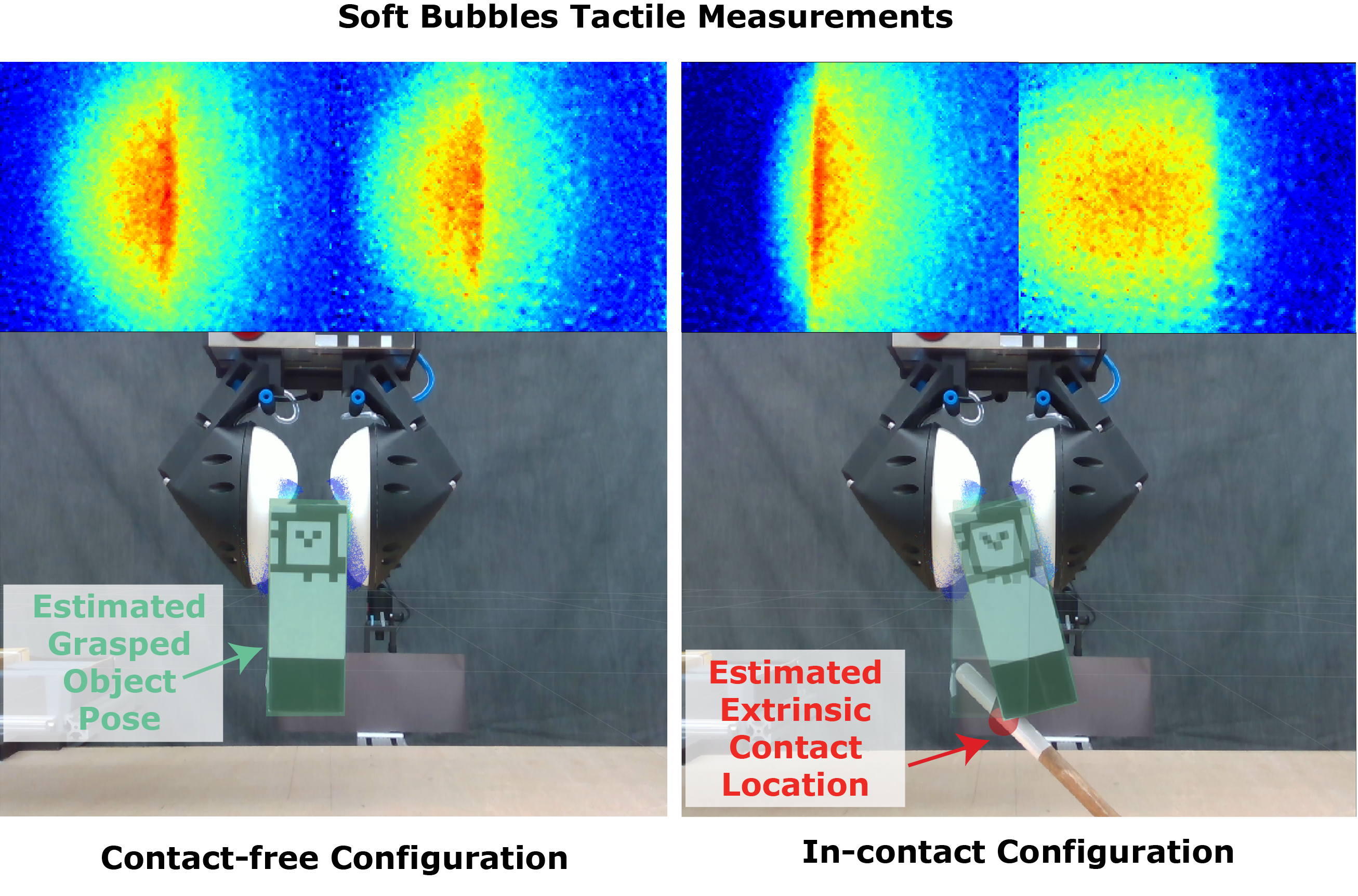}
    \caption{\textbf{Tactile State Estimation:} We show the state estimation for the grasped object for contact-free (left) and in-contact (right) configurations. We overlay the estimated object pose (turquoise) and the estimated contact location (red). On the in-contact configuration (right) we also overlay the contact-free configuration with a small alpha value. Note that the compliance of the sensors results in significant displacement between the contact-free and the in-contact configurations. 
    }
    \label{fig:state_estimation}
\end{figure}

To estimate the contact location $\graspedContactPoint$, we use the sensed wrench $\externalWrench$. There are two methods to recover this wrench that are interchangeable: 1) Use a wrist-mounted force-torque sensor to directly measure the wrench or 2) Use the wrench obtained via the passive compliance model (introduced in the next subsection). We assume that the contact is a point contact and to recover this point, we form the following program:
\begin{align*}
    & \min_{\graspedContactPoint, \vec{f}_c} [\externalWrench - \mat{J}^\top(\graspedContactPoint, \vec{x}_\text{go})\vec{f}_c ]^\top[\externalWrench - \mat{J}(\graspedContactPoint, \vec{x}_\text{go})\vec{f}_c ] \\
    & \text{s.t.} \quad \graspedContactPoint \in \mathcal S, \quad \vec{f}_c \in \mathcal{F}_c 
\end{align*}

where $\mat{J}(\graspedContactPoint, \vec{x}_\text{go})$ represents the Jacobian to the point $\graspedContactPoint$ on the surface of the grasped object $\mathcal S$, $\vec{f}_c$ is the contact force, and $\mathcal{F}_c$ is the friction cone. As \citet{contact_particle_filter} observe, this optimization program is generally non-convex and can be difficult to solve. However, by fixing $\graspedContactPoint$ and assuming that $\mathcal{F}_c$ is determined by Coulomb friction, the problem is then a convex QP. 
Then, to determine the contact point $\graspedContactPoint$ we can initialize different contact candidates $\graspedContactPointSub{,i} \in \mathcal S,\quad i=1,\dots,K$ and formulate a Contact Particle Filter (CPF) \cite{contact_particle_filter} to find the best contact that explains the sensed wrench. 
This optimization is easily parallelized across the faces of the polygon and can be solved efficiently \cite{multiscope}. This optimization can be further sped up by noting that surfaces likely to be in contact can be quickly narrowed down to just a few sides given priors on the object poses. Figure \ref{fig:state_estimation} illustrates our tactile state estimation running in real-time.



\begin{figure*}[h!]
    \centering
    \includegraphics[width=0.9\textwidth]{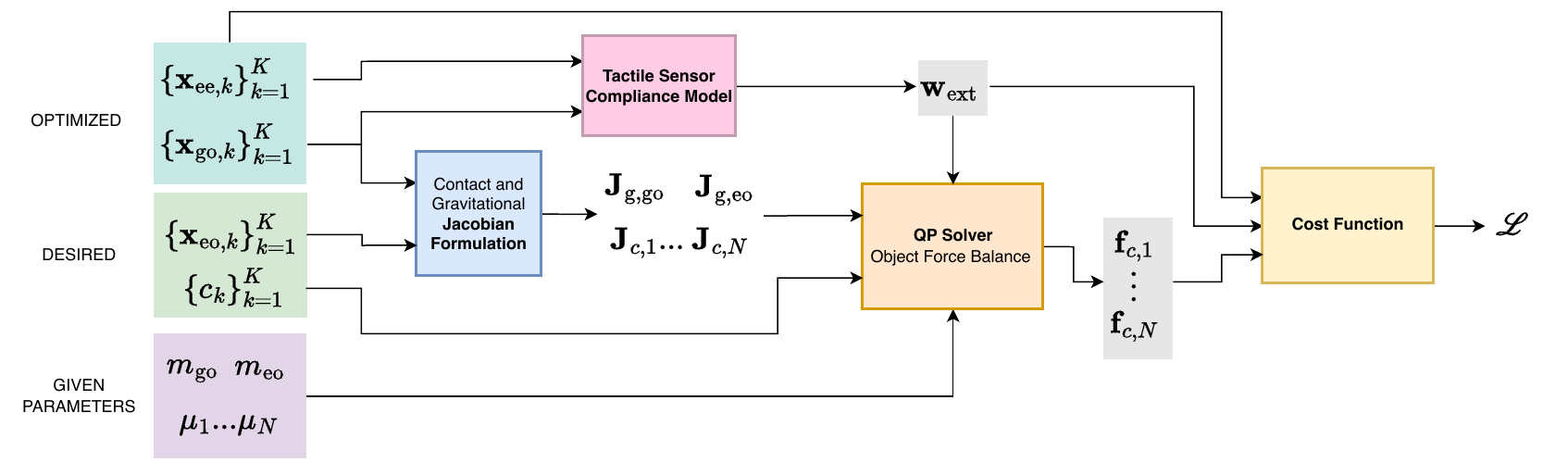}
    \caption{\textbf{Trajectory Optimization Overview:} Given a desired trajectory of the extrinsic object $\{\extrinsicObjectStateSub{,k}\}_{k=1}^K$ as well as the contact modes $\contactModeSeq$, our method optimizes the end-effector poses $\{\robotStateSub{,k}\}_{k=1}^K$ and graped object poses $\{\graspedObjectStateSub{,k}\}_{k=1}^K$. All blocks are differentiable, which enables the use of gradient-based optimization.}
    \label{fig:optimization_overview}
\end{figure*}

\subsection{Tactile Elasticity Model}
The goal of the tactile elasticity module is to model the force-deformation relationship of the Soft Bubbles tactile sensor for a grasped object. Here, we follow the SEED approach \cite{seed}, where they showed that the compliance of the Soft Bubbles sensors can be exploited to turn a force control problem into a position control. The key insight is to model the sensor elasticity as a series of linear elastic components.

The SEED model relates a measured grasped object displacement $\bold x_\text{ee}^\text{gocf} = \begin{bmatrix} x&y&\theta\end{bmatrix}^\top$ obtained using the method described in Section \ref{sec:tactile_state_estimation}, to a corresponded wrench $\bold w_\text{ext} = \begin{bmatrix} f_x&f_y&\tau\end{bmatrix}^\top$. Note that this displacement is measured in the object-sensor contact frame, denoted by $\text{gocf}$. This relationship is linear and is given by
\begin{equation}
\bold w_\text{ext} = \bold K  \bold x_\text{ee}^\text{gocf}
\label{eq:elasticity}
\end{equation}
where $\bold K \in \R^{3\times 3}, \bold K \succeq \bold 0$. At control time, to achieve a given desired and feasible external wrench $\hat{\bold w}_\text{ext}$ is equivalent to setting a relative displacement between the grasped object and the robot end-effector $\hat{\bold x}_\text{ee}^\text{gocf}= \bold K^{-1} \hat{\bold w}_\text{ext}$.

An
important consideration here is that the elasticity model is agnostic to whether or not this external wrench is achievable. However, not all configurations and wrenches are feasible since there must be an external contact on the grasped object that produces the robot-object configuration obeying the friction constraints. This motivates explicit reasoning over contact constraints as described in the next sections.

 \subsection{Extrinsic Contact Model}
The goal of the extrinsic contact model is to model the dynamics of the extrinsic object in contact with the grasped object and the environment plane. Given the grasped and extrinsic object poses, their intrinsic parameters such as the center of mass (CoM), mass $m$, geometries and friction properties $\mu$, and the contact locations, we want to solve for the contact forces $\bold f_\text{c,1},  \dots, \bold f_\text{c,N}, \quad \bold f_\text{c,i} =\begin{bmatrix}f_{c,i,n} & f_{c,i,t} \end{bmatrix}^\top \in \R^2$ that result in static equilibrium. Therefore, the contact forces  must be the solution to the following optimization problem:

\begin{mini*}|s|
{\bold f_\text{c,1},  \dots, \bold f_\text{c,N}}{\sum_{i=1}^N \frac{1}{2}\bold f_\text{c,i}^\top\bold U\bold f_\text{c,i} \hspace{11em} \mathrm{(P1)}}
{}{}
\addConstraint{}{\boldsymbol\tau_{g,\text{eo}} + \sum_{i=1}^N \bold J_\text{c,i}^\top\bold f_\text{c,i} = 0 }{}{}
\addConstraint{}{\bold w_\text{ext} + \boldsymbol\tau_{g,\text{go}} + \sum_{j\in \text{obj}} \bold J_\text{c,go,j}^\top\bold f_\text{c,j} = 0}{}{}
\addConstraint{}{0 \leq \mu_i f_{c,i,n} + f_{c,i,t} \perp \dot{s}_{c,i}^{+} \geq 0}{\quad}{i=1, \dots, N}
\addConstraint{}{0 \leq \mu_i f_{c,i,n} - f_{c,i,t} \perp \dot{s}_{c,i}^{-} \geq 0}{\quad}{i=1, \dots, N}
\end{mini*}

\noindent where $\boldsymbol \tau_{g, \cdot}$ are the gravitational terms and $\bold J_{c}$ are the contact Jacobians. They depend on the object CoM location and the object configuration. The first equality constraint expresses the force balance for the extrinsic object for static equilibrium. The second equality constraint is the force balance for the grasped object. Finally, the inequality complementary constraints are the contact friction constraints and model Coulomb friction. They express that contact forces must lie on the interior of the friction cone when contact sticks, and on the friction cone boundary when contact slips.
Note that $0 \leq a \perp b \geq 0$ implies that $a\geq0, b\geq0$ and $a \cdot b=0$. 

To account for the different complementary constraints, for each of the different contact modes, we define an instance of $\mathrm{(P1)}$ with the complementary constraints determined. Therefore, given the contact mode, the complementary constraints are turned into equality or inequality constraints. We model the contact with the ground plane as two-point contact located at the edges of the contact face. When the contact is a line contact, both points are active. On the other hand, when there is a single contact point, only the contact vertex is active.


The contact model for extrinsic object manipulation defined in $\mathrm{(P1)}$ can be simplified to express the contact dynamics of the grasped object directly in contact with the static environment. Note that this is equivalent to assuming that the extrinsic object is rigidly attached to the environment. The resultant contact model is of the form: 

\begin{mini*}|s|
{\bold f_\text{c,1},  \dots, \bold f_\text{c,N}}{\sum_{i=1}^N \frac{1}{2}\bold f_\text{c,i}^\top\bold U\bold f_\text{c,i} \hspace{11em} \mathrm{(P2)}}
{}{}
\addConstraint{}{\bold w_\text{ext} + \boldsymbol\tau_{g,\text{go}} + \sum_{i=1}^N \bold J_\text{c,i}^\top\bold f_\text{c,i} = 0 }{}{}
\addConstraint{}{0 \leq \mu_i f_{c,i,n} + f_{c,i,t} \perp \dot{s}_{c,i}^{+} \geq 0}{\quad}{i=1, \dots, N}
\addConstraint{}{0 \leq \mu_i f_{c,i,n} - f_{c,i,t} \perp \dot{s}_{c,i}^{-} \geq 0}{\quad}{i=1, \dots, N}
\end{mini*}
Note that $\mathrm{(P2)}$ has one equality constraint -- the force balance of the grasped object in contact with the static environment.

\subsection{Extrinsic Contact Trajectory Optimization}

The goal of the controller is to generate a trajectory of end-effector and grasped object poses that results in the desired motions of the extrinsic object and transmitted forces. However, as discussed previously, not all motions are feasible since they must satisfy geometric and contact constraints, maintain equilibrium, and account for sensor compliance. The key contribution of our method is to formulate the contact trajectory optimization precisely to address these requirements while also being amenable to gradient-based optimization and capable of producing a variety of ``manipulation skills'' described by the prescribed contact modes. Our method integrates the 3 previous components into a unified framework for perception (object and extrinsic contact detection), planning (solving $\mathrm{(P1)}$ for extrinsic object manipulation or $\mathrm{(P2)}$ for grasped object manipulation together with the compliance model), and feedback controllers to stabilize planned trajectories (see Sec. \ref{sec:exp_prehensile_control}).

In general, the trajectory optimization problem is non-convex since the force balance on the extrinsic and grasped object in $\mathrm{(P1)}$ depends on Jacobians and gravitational term. These terms depend on the configuration of the extrinsic and the grasped object states $\extrinsicObjectState, \graspedObjectState$ through sinusoidal functions, resulting in a non-convex problem. However, we observe that given the end-effector pose $\robotState$ and the object configurations $\graspedObjectState$ and $\extrinsicObjectState$, this optimization problem is a Quadratic Program (QP) since the Jacobians and gravitational terms are determined. Therefore, we construct the Jacobians out of the optimization program so they are just an optimization parameter and they do not depend on the optimization variables. Moreover, since the Jacobian formation is differentiable, inspired by OptNet \cite{optnet}, we can propagate the gradients through the QP and employ gradient-based optimization to iteratively update the poses of the end-effector $\{\robotStateSub{,k}\}_{k=1}^K$ and the grasped object trajectory $\{\graspedObjectStateSub{,k}\}_{k=1}^K$.  


To solve this, our algorithm performs different computations, which are illustrated in figure \ref{fig:optimization_overview}. In particular, for each step of the trajectory do:

\begin{enumerate}
    \item Given $\extrinsicObjectState, \graspedObjectState$, create the Jacobians $\bold J_\text{c,1},\dots, \bold J_\text{c,N}$ for each contact point as well as the gravitational Jacobians $\bold J_\text{g, eo}, \bold J_\text{g, go}$.
    \item Given the grasped object state $\graspedObjectState$ and the robot state $\robotState$ use the tactile compliance model to compute the external wrench applied to the grasped object $\bold w_\text{ext}$.
    \item Given the Jacobians, the external wrench $\bold w_\text{ext}$, the object masses, friction coefficients for each contact, and the contact modes $c$ formulate the QP given by $\mathrm{(P1)}$. The solution to the program is the contact forces $\bold f_\text{c,1}, \dots, \bold f_\text{c,N}$.
    \item Given the object and robot poses, the external wrench, and the contact forces compute the loss function $\mathcal L$ and backpropagate the gradients through the different blocks to update the grasped object and robot trajectories.
\end{enumerate}

One limitation of this approach is that during the optimization the poses may be updated to unfeasible values, i.e. the resultant QP is not solvable. When this happens, there are no gradients since the QP is not feasible. To account for this, we relax the grasped-extrinsic object contact friction cone constraints and replace them with a cost term that pushes these contact forces to the contact cone.

The resultant cost function is defined as follows:
\begin{equation*}
\mathcal L = \mathcal L_\text{cone} +  \mathcal L_\text{smooth} +  \mathcal L_\text{contact force} +  \mathcal L_\text{penetration} 
\end{equation*}
where:

\begin{itemize}
    \item Cone Loss ($\mathcal L_\text{cone}$): Incentivizes the contact forces between the grasped object and the extrinsic object to be within the cone boundaries. It is defined as:
        \begin{equation*}
        \mathcal L_\text{cone} = \frac{1}{K}\sum_{k=1}^K (1-\bold \langle \frac{\bold f_\text{c,obj,k}}{\|\bold f_\text{c,obj,k}\|}, \begin{bmatrix}0\\1\end{bmatrix}\rangle)^2 
        \end{equation*}
    \item Smooth Loss ($\mathcal L_\text{smooth}$): This loss encourages the grasped object trajectory to be smooth by preventing the sequence of poses from having large changes in magnitude.
        \begin{equation*}
         \mathcal L_\text{smooth}  = \frac{1}{K-1}\sum_{k=1}^{K-1} \|\graspedObjectStateSub{,k+1} - \graspedObjectStateSub{,k}\|^2
        \end{equation*}
    \item Contact Force Loss ($\mathcal L_\text{contact force}$): The contact force loss incentivizes the contact forces to be close to a desired contact force.
        \begin{equation*}
         \mathcal L_\text{contact force} = \frac{1}{K}\sum_{k=1}^K \| \bold f_\text{c,obj,k} - \bold f_\text{c, obj,k}^\text{des}  \|^2
        \end{equation*}
    \item Penetration Loss ($\mathcal L_\text{contact force}$): Limits the range of the admissible relative angles to avoid penetration. We use the log-barrier function to enforce this constraint.
        \begin{equation*}
         \mathcal L_\text{penetration} =  \frac{1}{K}\sum_{k=1}^K \|-\frac{1}{10}\log(-\phi_k)\|^2
        \end{equation*}
        
\end{itemize}



Our implementation is based on \texttt{cvxpylayers} \cite{cvxpylayers} to differentiate over the QP and \texttt{pytorch} autograd for gradient computation and update the tensors.

\section{Experiments and Results}


\begin{figure*}[h!]
    \centering
    \includegraphics[width=\textwidth]
    {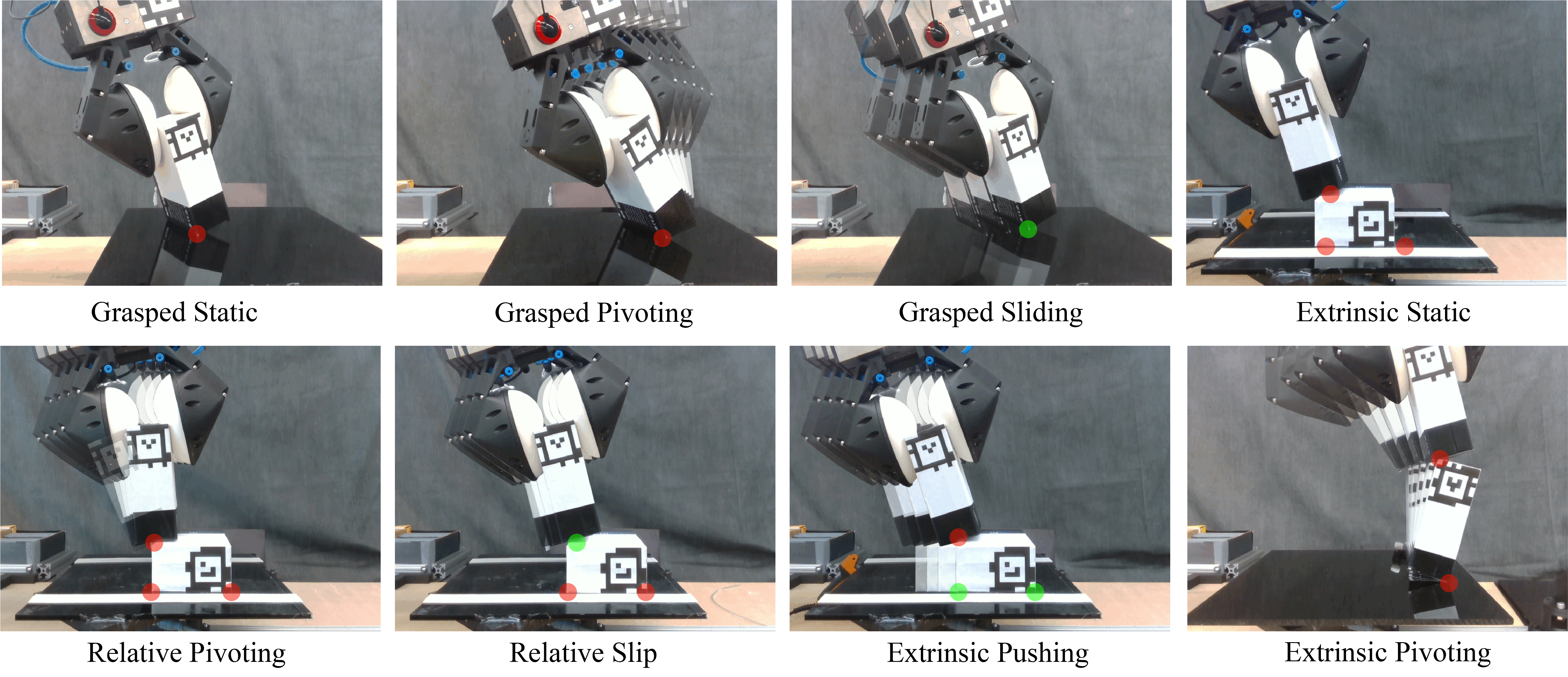}
    \caption{\textbf{In-Contact Manipulation Skills:} We show our framework on a diverse set of in-contact skills. The first 3 are for manipulating a grasped object in contact with the environment. The remaining 5 are for non-prehensile manipulation. We display the sticking contact points in red and the slipping contacts in green.}
    \label{fig:primitives}
\end{figure*}

\subsection{Tactile Elasticity Characterization}
\label{sec:experiments_tactile_elasticity_char}
\begin{figure*}[h!]
    \centering
    \includegraphics[width=\textwidth]
    {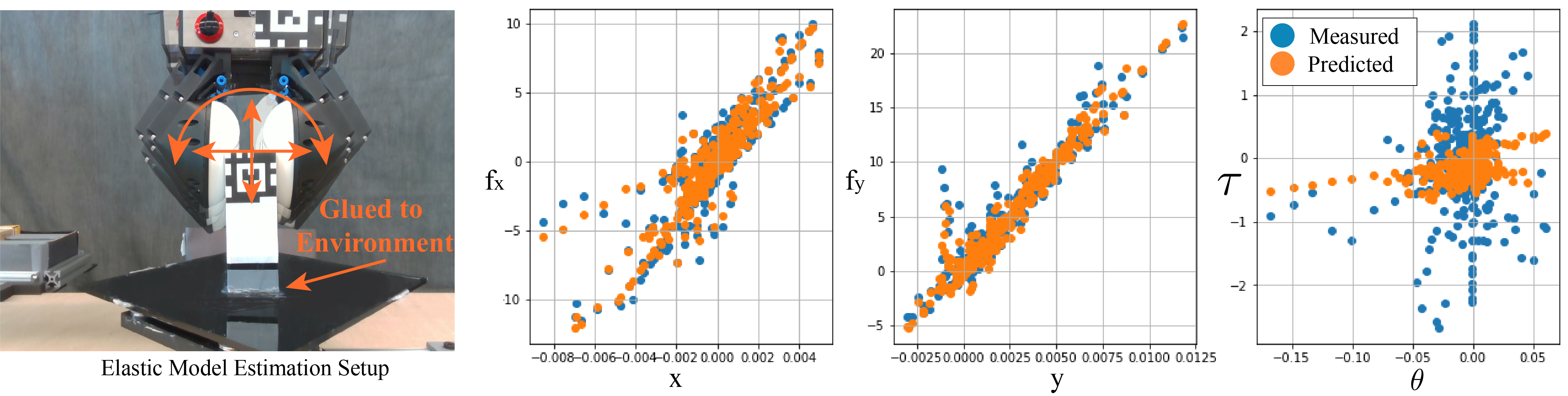}
    \caption{\textbf{Tactile Elasticity Identification:} (Left) During the data collection, the robot grasps an object rigidly attached to the environment while moving the end effector. (Right) We plot the observed data displacements vs. measured wrenches (on blue) and the fitted model predictions (orange).}
    \label{fig:elasticity_estimation}
\end{figure*}
To characterize the compliance of the tactile sensors, the robot collects data by attempting to manipulate an object fixed to the environment, Fig.~\ref{fig:elasticity_estimation}. In more detail, the robot moves its end-effector to sampled relative motions with respect to the fixed object while maintaining contact with the membranes. We use the Gamma ATI force-torque sensor mounted on the wrist of the robot to collect ground truth force values and grasped object poses. The grasped object poses are obtained using the AprilTag fiducial marker on the object and a calibrated camera. 
We collect a total of 200 force-displacement samples under random motions and motions along the principal axis. 
Given the collected data, we perform regression to obtain the elasticity matrix $\bold K\succeq \bold 0$ that best models equation \ref{eq:elasticity}. We constrain $\bold K$ to be PSD by forcing it to be the $\bold K = \bold L^\top \bold L$. We use \texttt{pytorch} to optimize this matrix.
The resultant value is:
\begin{equation*}
\bold K_\text{bubbles} = \begin{bmatrix} 2208.96 & -143.31 &  399.86\\
 -143.31 & 1875.92 & -313.57\\
399.86 & -313.57 &   1608.49 \end{bmatrix}
\end{equation*}
Fig.~\ref{fig:elasticity_estimation} (right) illustrates the model fit to the data. We note that opting for a dense matrix enables modeling of the cross-interaction between the 3 degrees of freedom.

\subsection{Friction Model Identification}
\begin{figure}[h!]
    \centering
    \includegraphics[width=0.5\textwidth]
    {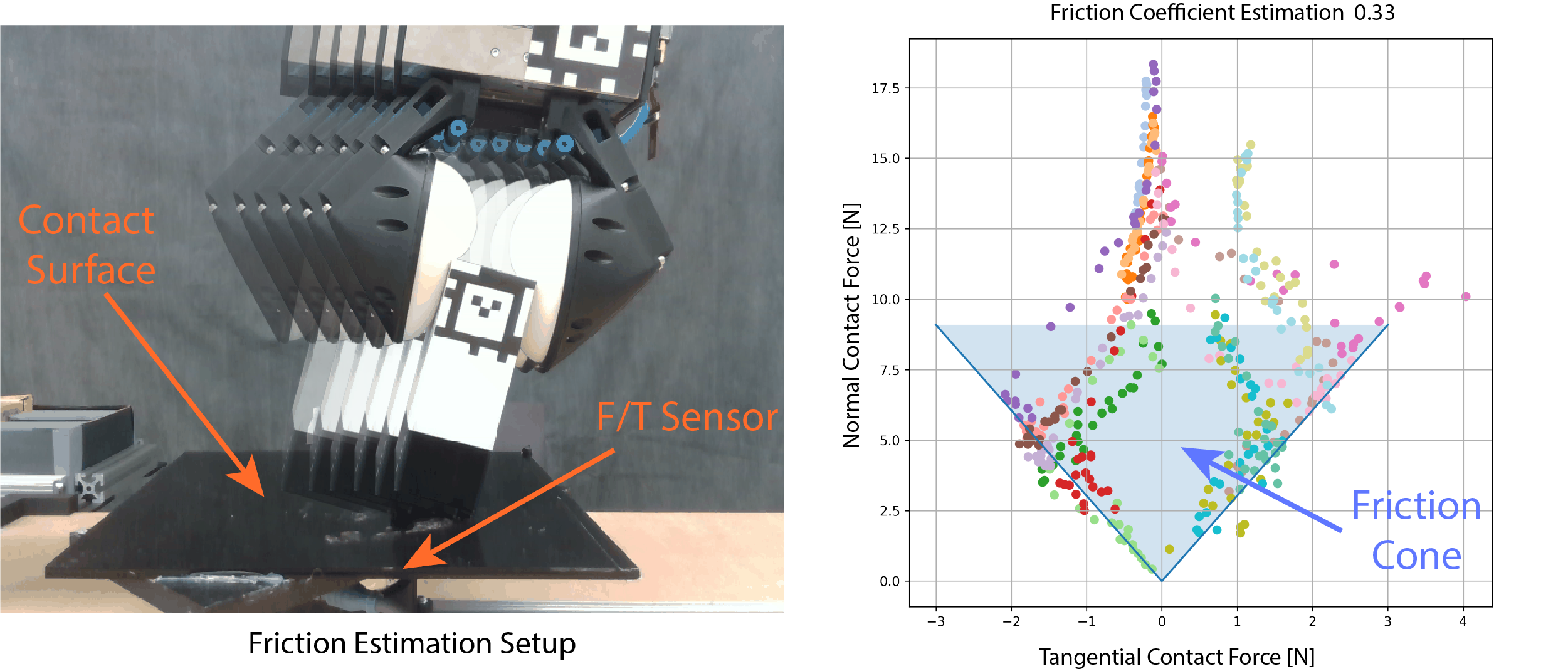}
    \caption{\textbf{Fiction Cone Estimation:} (Left) During the friction coefficient identification, the robot moves while maintaining the object in contact with the contact surface while recording the contact forces with the force-torque sensor attached under the surface. (Right) We visualize the contact forces for each of the different trajectories performed. The samples lie within a cone that defines the friction coefficient to be $\mu=0.33$}
    \label{fig:friction_model}
\end{figure}

In this section, we describe the methods used to estimate the friction coefficients $\mu$ that describe the contact interactions between objects and object-to-environment. Fig.~\ref{fig:friction_model} (left) illustrates the setup used. For each contact surface pair, we use the robot to grasp the object and execute random in-contact trajectories. During such trajectories, we record the contact forces $\bold f_c = \begin{bmatrix}f_t & f_n \end{bmatrix}^\top$ measured by the F/T sensor attached under the manipulation plane. For object-object interactions, we attach the object to the scene similar to the setup described in section \ref{sec:experiments_tactile_elasticity_char} and have the robot perform object-object interactions. Fig.~\ref{fig:friction_model} (right) for object-plane interaction. Observe that the force measurements lie within a cone, validating the Coulomb friction assumption. For the shown case, the estimated friction coefficient is $\mu=0.33$.

\subsection{Prehensile Object Extrinsic Contact Control} \label{sec:exp_prehensile_control}
The goal of this section is to verify the proposed contact formulation and sensor compliance model. To this end, we consider the problem of manipulating a grasped object while in contact with the environment. Specifically, we consider 3 distinct in-contact manipulation skills. These skills are:
\begin{itemize}
    \item \textit{Static in Contact Manipulation}: The goal is to maintain the grasped object at a fixed in-contact configuration while achieving a desired contact force. In this instance, the contacts between the object and the environment must be sticking, i.e. $\bold f_{c,i} \in\text{int}\ \mathcal F_{c,i}$.
    \item \textit{Grasped Object Pivoting}: The goal is to change the orientation of the grasped object while maintaining a desired contact force with the environment. 
    \item \textit{Grasped Object Sliding}: In this case, we want the object trajectory to slip w.r.t to the environment, i.e. $\bold f_{c,i} \in \partial\mathcal F_{c,i}$ while transmitting a desired contact force to the environment.
\end{itemize}
We chose to demonstrate these skills first since they are simpler instances of the extrinsic manipulation problem and enable us to demonstrate the basic elements of our proposed method. In Sec.~\ref{sec:exp_non_prehensile_control} we consider the general case.
To execute the off-line optimized trajectory for these skills, we evaluate 4 different controllers. These controllers are:
\begin{enumerate}
    \item \textit{Open-loop Controller}: This controller executes the offline optimized trajectory.
    \item \textit{Closed-loop Controller}: Given the optimized trajectory, this model uses the current measurements for the grasped object pose $\graspedObjectState$ and measured wrench $\bold w_\text{ext}$ to compensate the given trajectory.
    \item \textit{Closed-loop PI Controller}: This controller extends the previous one by limiting the update using proportional and integrative terms given by $k_\text{p}$ and $k_\text{i}$ constants.
    \item \textit{Rigid Sensor Controller}: This controller assumes that the grasped object is rigidly attached to the robot, i.e. the grasped object pose is constant w.r.t. to the robot end-effector.
\end{enumerate}

In order to evaluate how well a skill is performed, we measure the error between the desired object pose and the transmitted wrenches to the environment and the ground truth measurements obtained using the Gamma ATI force-torque sensor and Apriltag markers along the trajectory. Table \ref{tab:grasped_control_eval} summarizes the results. Our experiments show that the closed-loop controllers achieve superior performance tracking the desired trajectories than the other tested control approaches. Hence, we decide to use the PI closed-loop controller for the rest of the experiments. 

\begin{table*}[h!]
\vspace{-5pt}
\centering
\begin{tabular}{lcccccccccccc}
\toprule
\multicolumn{1}{l}{\multirow{3}{*}{Controller}} & \multicolumn{4}{c}{Grasped Static} & \multicolumn{4}{c}{Grasped Pivoting} & \multicolumn{4}{c}{Grasped Sliding} \\ \cmidrule(lr){2-5}\cmidrule(lr){6-9}\cmidrule(lr){10-13}
\multicolumn{1}{c}{} & \multicolumn{2}{c}{Wrench Error [N]} & \multicolumn{2}{c}{Pose Error [mm]} &  \multicolumn{2}{c}{Wrench Error [N]}  & \multicolumn{2}{c}{Pose Error [mm]} &\multicolumn{2}{c}{Wrench Error [N]} & \multicolumn{2}{c}{Pose Error [mm]} \\ 
\multicolumn{1}{c}{} & \multicolumn{1}{c}{Mean $\downarrow$} & \multicolumn{1}{c}{Std $\downarrow$} & \multicolumn{1}{c}{Mean $\downarrow$} & Std $\downarrow$ & \multicolumn{1}{c}{Mean$\downarrow$} & \multicolumn{1}{c}{Std$\downarrow$} & \multicolumn{1}{c}{Mean$\downarrow$} & Std $\downarrow$ & \multicolumn{1}{c}{Mean$\downarrow$} & \multicolumn{1}{c}{Std$\downarrow$} & \multicolumn{1}{c}{Mean$\downarrow$} & Std $\downarrow$\\ \midrule

\begin{tabular}[c]{@{}l@{}}Open-loop\end{tabular} & 0.75 & 0.27 & 1.35 & 0.23 & 1.15 & 0.08 & 1.86 & 0.23 & 0.76 & 0.21 & 1.94 & 0.84  \\

\begin{tabular}[c]{@{}l@{}}Closed-loop\end{tabular} & 0.45 & 0.18 & 0.94 & 0.33 & 0.50 & 0.09 & 0.52 & 0.12 & 0.35 & 0.16 & 0.569 & 0.31  \\ 

\begin{tabular}[c]{@{}l@{}}Closed-loop PI\end{tabular}& 0.51 & 0.19 & 0.85 & 0.23 & 0.64 & 0.09 & 0.48 & 0.13 & 0.22 & 0.10 & 0.478 & 0.26  \\

\begin{tabular}[c]{@{}l@{}}Rigid Sensor\end{tabular} & 1.74 & 0.22 & 1.97 & 0.87 & 1.9 & 0.09 & 0.39 & 0.27 & 0.79 & 0.18 & 0.78 & 0.93  \\

\bottomrule
\end{tabular}
\vspace*{3pt}
\caption{\textbf{Grasped Object Manipulation Controller Evaluation:} We benchmark different control methods for tracking a desired pose and transmitted poses. We aggregate the orientation and toques using the radius of gyration of the grasped object.}
\label{tab:grasped_control_eval}
\end{table*}

\begin{table}[h!]
\vspace{-5pt}
\centering
\begin{tabular}{lcccc}
\toprule
\multicolumn{1}{l}{\multirow{3}{*}{Manipulation Skill}} & \multicolumn{2}{c}{Grasped Object} & \multicolumn{2}{c}{Extrinsic Object}  \\ 
\multicolumn{1}{c}{} & \multicolumn{2}{c}{Pose Error [mm]} & \multicolumn{2}{c}{Pose Error [mm]}  \\ \cmidrule(lr){2-3}\cmidrule(lr){4-5}
\multicolumn{1}{c}{} & \multicolumn{1}{c}{Mean $\downarrow$} & \multicolumn{1}{c}{Std $\downarrow$} & \multicolumn{1}{c}{Mean $\downarrow$} & Std $\downarrow$ \\ \midrule

\begin{tabular}[c]{@{}l@{}}Static Contact (\texttt{hbox}) \end{tabular} & 0.78 & 0.55 & 0.22 & 0.12  \\

\begin{tabular}[c]{@{}l@{}}Relative Pivoting (\texttt{hbox}) \end{tabular} & 1.26 & 0.98 & 0.31 & 0.32  \\ 

\begin{tabular}[c]{@{}l@{}}Relative Sliding (\texttt{hbox}) \end{tabular} &  1.86 & 1.33  & 0.36 & 0.47 \\

\begin{tabular}[c]{@{}l@{}}Extrinsic Pushing (\texttt{hbox})\end{tabular} & 2.87 & 2.21  & 2.38 & 1.89  \\

\bottomrule
\end{tabular}
\vspace*{3pt}
\caption{\textbf{Extrinsic Object Manipulation Skills Evaluation:} We evaluate different manipulation skills on the trajectory error of the grasped object and the extrinsic object poses. We aggregate orientations using the radius of gyration for each object geometry.}
\label{tab:extrinsic_skills_eval}
\end{table}

\subsection{Non-Prehensile Object Manipulation via Extrinsic Contact Control} \label{sec:exp_non_prehensile_control}

In this section, we evaluate our proposed method for manipulating the extrinsic object with the grasped object. Similar to Sec.~\ref{sec:exp_prehensile_control}, we propose a set of manipulation skills to benchmark our proposed method for non-prehensile in-contact object manipulation.
These manipulation skills, illustrated in Fig.~\ref{fig:primitives}, are:
\begin{enumerate}
    \item \textit{Static in Contact Manipulation}: The goal is to maintain the grasped object at a fixed in-contact configuration with the extrinsic object while achieving a desired contact force. In this instance, the contacts between the objects and object environment must be sticking, i.e. $\bold f_{c,i} \in\text{int}\ \mathcal F_{c,i}$.
    \item \textit{Relative Pivoting}: The goal is to change the orientation of the grasped object while maintaining a desired contact with the extrinsic object while not allowing the extrinsic object to slip w.r.t. the environment. 
    \item \textit{Relative Sliding}: The grasped object trajectory must slip w.r.t to the extrinsic object, i.e. $\bold f_{c,i} \in \partial\mathcal F_{c,i}$ while transmitting a desired contact force but maintaining the sticking contact between the extrinsic object and environment.
    \item \textit{Extrinsic Pushing}: The goal of extrinsic pushing is to use the grasped object as a means to push the extrinsic object. The desired contact mode is sticking contact between the grasped and extrinsic objects contacts, while the contact between the extrinsic object and the environment must be slipping.
    \item \textit{Extrinsic Pivoting}: In this case, we want the object trajectory to slip w.r.t to the environment, i.e. $\bold f_{c,i} \in \partial\mathcal F_{c,i}$ while transmitting a desired contact force to the environment. This manipulation mode is the hardest of all since it requires the extrinsic object to be on non-stable configurations.
\end{enumerate}

\begin{figure}[h!]
    \centering
    \includegraphics[width=0.5\textwidth]
    {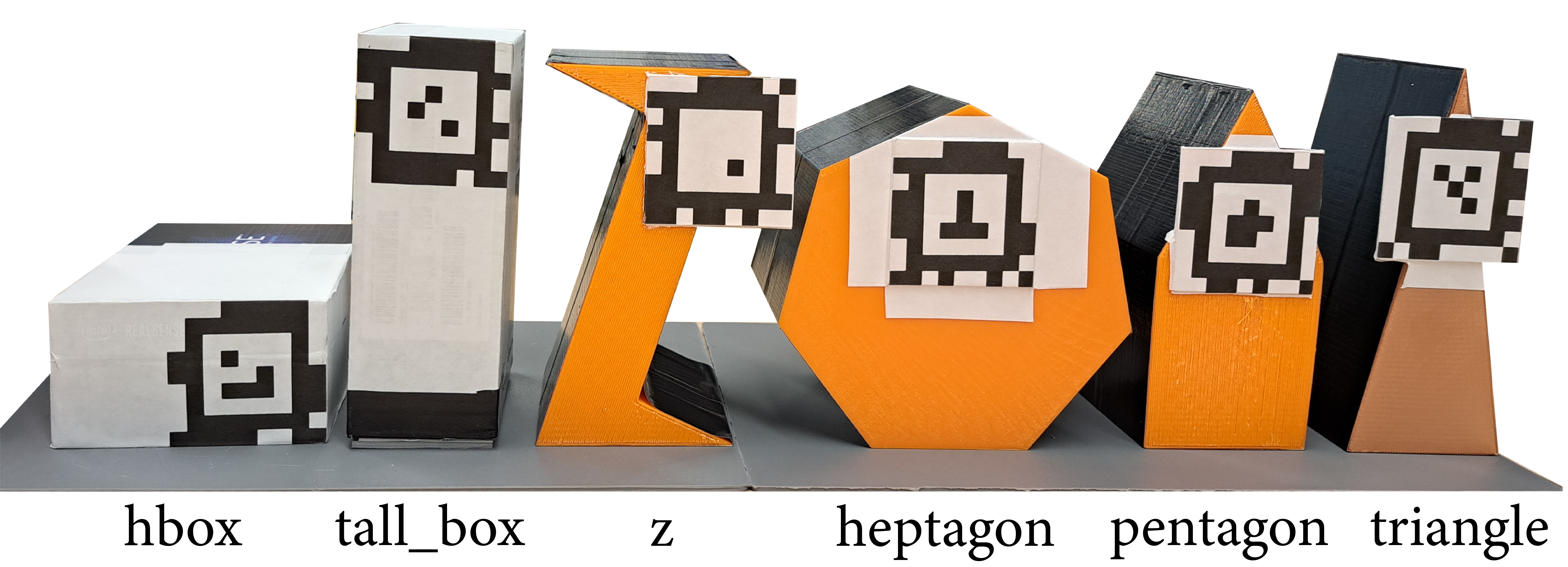}
    \caption{\textbf{Manipulation Shapes Shapes:} We test our method on a diverse set of planar polygonal shapes. Note that our approach can handle convex as well as non-convex geometries. \texttt{hbox} is used to evaluate extrinsic manipulation skills 1-4 while the remaining geometries are used to evaluate extrinsic pivoting.}
    \label{fig:shapes}
\end{figure}

We evaluate these manipulation primitives in a diverse set of polygonal object geometries, shown in figure \ref{fig:shapes}. We use the first object (\texttt{hbox}) for benchmarking primitives 1-4 while using the remaining more diverse and complex objects for extrinsic pivoting since it is the hardest primitive. As the grasped object, we use the same object as Sec.~\ref{sec:exp_prehensile_control} for prehensile control, which has the same geometry as the \texttt{tall\_box} object.

For each primitive 1-4, we perform 3 trajectories of 20 steps. We measure the pose errors between the desired trajectories and the measured trajectories for the grasped object as well as the extrinsic object. Table \ref{tab:extrinsic_skills_eval} summarizes the results. We observe that we obtain sub-millimeter errors when the pose is static while getting higher errors when larger motions of the objects are required.

To evaluate the extrinsic pivoting performance, for each of the objects we perform 3 pivoting trajectories of 40 steps each. The desired trajectories are optimized so the contact force between objects is 3N. Table \ref{tab:extrinsic_pivoting_results} summarizes the evaluation for extrinsic pivoting. We report the mean absolute error for each of the wrench and pose components.  We observe that we achieve errors below 1N for force and in the order of a millimeter accuracy for the pose tracking error. These results demonstrate the efficacy of our approach and the successful completion of the manipulation skills. We do observe that the heptagon pose error is higher than the other geometries and we hypothesize that this is due to the fact that this object has the largest area.

\begin{table*}[h!]
\vspace{-5pt}
\centering
\resizebox{\textwidth}{!}{%
\begin{tabular}{lcccccccccccccc}
\toprule
\multicolumn{1}{l}{\multirow{3}{*}{Extrinsic Object}} & \multicolumn{7}{c}{Wrench Errors} & \multicolumn{7}{c}{Pose Errors} \\ \cmidrule(lr){2-8}\cmidrule(lr){9-15}
\multicolumn{1}{c}{} & \multicolumn{2}{c}{$f_x$ Error [N]} & \multicolumn{2}{c}{$f_y$ Error [N]} &  \multicolumn{2}{c}{$\tau$  Error [Nmm]} & \multicolumn{1}{c}{Summary}  & \multicolumn{2}{c}{$x$ Error [mm]} &\multicolumn{2}{c}{$y$ Error [mm]} & \multicolumn{2}{c}{$\theta$ Error [deg]} & \multicolumn{1}{c}{Summary} \\ 
\multicolumn{1}{c}{} & \multicolumn{1}{c}{Mean $\downarrow$} & \multicolumn{1}{c}{Std $\downarrow$} & \multicolumn{1}{c}{Mean $\downarrow$} & Std $\downarrow$ & \multicolumn{1}{c}{Mean$\downarrow$} & \multicolumn{1}{c}{Std$\downarrow$} &  [N] & \multicolumn{1}{c}{Mean$\downarrow$} & Std $\downarrow$ & \multicolumn{1}{c}{Mean$\downarrow$} & \multicolumn{1}{c}{Std$\downarrow$} & \multicolumn{1}{c}{Mean$\downarrow$} & Std $\downarrow$ & [mm]\\ \midrule

\texttt{tall\_box} & 0.63 & 0.53 & 0.77 & 0.66 & 86.9 & 53.47 & 2.18
& 0.36 & 0.26 & 0.72 & 0.68 & 0.68 & 0.40 & 0.805 \\ 

\texttt{z} & 0.65 & 0.52 & 0.87 & 0.71 & 63.4 & 52.7 & 1.74
& 0.46 & 0.31 & 0.68 & 0.43 & 1.29 & 1.11 & 0.821 \\ 

\texttt{heptagon} & 0.61 & 0.51 & 0.98 & 0.78 & 24.8 & 19.96 & 1.27
& 2.02 & 1.45 & 0.56 & 0.41 & 1.73 & 1.85 & 2.096 \\

\texttt{pentagon} & 0.23 & 0.25 & 0.34 & 0.15 & 26.5 & 13.6 & 0.75
&1.25 & 0.63 & 0.67 & 0.53 & 3.34 & 1.69 & 1.419  \\

\texttt{triangle}& 0.48 & 0.39 & 0.57 & 0.59 & 99.7 & 66.1 & 2.82
& 0.74 & 0.42 & 0.67 & 0.48 & 1.41 & 1.22 & 0.998 \\

\bottomrule
\end{tabular}
}
\vspace*{3pt}
\caption{\textbf{Extrinsic Pivoting Trajectory Evaluation:} We measure the absolute error between the achieved and desired wrenches and poses for the pivoting trajectories. The Summary metric integrates wrenches (forces and torques) and poses (positions and angles) into a unified value using object geometry via the radius of gyration}. Our method achieves good tracking both for poses and wrenches.
\label{tab:extrinsic_pivoting_results}
\end{table*}


\begin{table*}[h!]
\vspace{-5pt}
\centering
\begin{tabular}{lcccccccccccc}
\toprule
\multicolumn{3}{c}{\multirow{1}{*}{Optimization Approach}} & \multicolumn{4}{c}{Bubble Pivoting} & \multicolumn{4}{c}{Gelslim Pivoting} & \multicolumn{2}{c}{\multirow{2}{*}{Computation Time [s]}}  \\
\cmidrule(lr){1-3}\cmidrule(lr){4-7} \cmidrule(lr){8-11} 
\multicolumn{1}{c}{\multirow{2}{*}{Name}}& \multirow{2}{*}{\#iters} & \multirow{2}{*}{\#samples} & \multicolumn{2}{c}{Pose Error [mm]} & \multicolumn{2}{c}{Wrench Error [N]} & \multicolumn{2}{c}{Pose Error [mm]} & \multicolumn{2}{c}{Wrench Error [N]} &   \\ 
\cmidrule(lr){4-5} \cmidrule(lr){6-7} \cmidrule(lr){8-9} \cmidrule(lr){10-11}  \cmidrule(lr){12-13}
\multicolumn{3}{c}{} & \multicolumn{1}{c}{Mean $\downarrow$} & \multicolumn{1}{c}{Std $\downarrow$} & \multicolumn{1}{c}{Mean $\downarrow$} & Std $\downarrow$ & \multicolumn{1}{c}{Mean $\downarrow$} & Std $\downarrow$ & \multicolumn{1}{c}{Mean $\downarrow$} & Std $\downarrow$ & \multicolumn{1}{c}{Mean $\downarrow$} & Std $\downarrow$ \\ \midrule
\texttt{Grad 100} (ours) & 100 & 1 & 2.45 & 0.48 & 0.43 & 0.35 & 2.73 & 2.16 & 7.01 & 2.13 & 7.19 & 0.14  \\
\texttt{MPPI 100} & 1 & 100 & 3.49 & 4.04 & 0.56 & 0.266 & 17.06 & 13.11 & 7.97 & 2.82 & 7.81 & 0.39\\
\texttt{MPPI 1000} & 10 & 100 & 2.79 & 3.32 & 0.77 & 0.52 & 9.18 & 9.86 & 6.58 & 3.04 & 75.6 & 5.24 \\
\texttt{iCEM 100} & 10 & 10 & 3.95 & 2.69 & 0.45 & 0.22 & 19.5 & 17.69 & 8.12 & 2.31 & 10.70 & 0.11 \\
\texttt{iCEM 1000} & 50 & 20 & 2.95 & 2.81 & 0.505 & 0.27 & 9.39 & 4.81 & 8.25 & 4.15 & 104.66 & 0.67 \\
\bottomrule
\end{tabular}
\vspace*{3pt}
\caption{\textbf{Trajectory Optimization Algorithm Benchmark} We evaluate different trajectory optimization methods for the Pivoting Pentagon task. Gradient-based optimization results in smoother trajectories, more amenable for quasi-static manipulation requiring less computation time.}
\label{tab:emto_benchmark}
\end{table*}

\subsection{Extrinsic Trajectory Optimization Benchmark} \label{sec:optimization benchmark}
To benchmark our method, we compare it with two trajectory optimization algorithms: 1) Model-Predictive Path Integral (MPPI) \cite{mppi} and 2) Improved Cross-Entropy Method (iCEM) \cite{icem}. Unlike our method, which utilizes gradient-based optimization, these methods employ sample-based optimization techniques.
We specifically evaluate the extrinsic pivoting skill, as it represents the most complex among the proposed skills. Furthermore, our evaluation focuses solely on the \texttt{pentagon} object, as the results on Table \ref{tab:extrinsic_pivoting_results} indicate that it poses a challenging and interesting shape.
To ensure a fair comparison with the baseline methods, we evaluate two different versions of each: one with 100 QP queries and another with 1000 queries. Our approach, labeled as \texttt{Grad 100}, requires 100 quadratic programming (QP) queries, mirroring the query count of \texttt{MPPI 100} and \texttt{iCEM 100}. It's worth noting that \texttt{MPPI 1000} and \texttt{iCEM 1000} necessitate ten times more QP queries. Additionally, it's important to highlight that iCEM typically requires more samples than MPPI, but iCEM requires more iterations to converge to a solution.
For each method, we perform 5 offline trajectory optimizations with a horizon of 40 steps. The results are stored for subsequent execution by the robot.
We evaluate the execution of the optimized trajectories using Soft Bubbles and measure pose and wrench errors. Table \ref{tab:emto_benchmark} provides a summary of the results. Our experiments indicate that trajectories generated by sample-based methods exhibit more noise, particularly when fewer query points are used.
We also report the computation time across the optimization. Note that our method, \texttt{MPPI 100}, and \texttt{iCEM 100} have comparable computation times since they have the same number of queries to the QP problem. As expected \texttt{MPPI 1000} and \texttt{iCEM 1000} require about 10 more times to optimize the trajectory since they also require 10 more times queries to the QP problems.


We further evaluate our method for a more rigid tactile sensor such as the Gelslims \cite{gelslim3}. 
The main requirement for our method is adjusting the tactile elasticity model. We adhere to the process outlined in Section \ref{sec:experiments_tactile_elasticity_char} to characterize their elasticity model. The fitted model is
\begin{equation*}
\bold K_\text{gelslim} = \begin{bmatrix} 5060.26 & -1261.90 &  -158.71\\
 -1261.90 & 9998.16 & 334.56\\
-158.71 & 334.56 & 37.6311 \end{bmatrix}
\end{equation*}

It is worth noting that the Gelslims sensors are approximately five times stiffer than the Soft Bubbles, as observed in previous experiments \cite{manipulation_via_membranes}.
We conducted a repeated evaluation for each of the benchmarked optimization methods using the previously saved optimized trajectories. The results of this evaluation are summarized in Table \ref{tab:emto_benchmark}.
In comparison to the results obtained for the Soft Bubbles, we observed similar outcomes for the Gelslim sensors. However, it is important to highlight that the sample-based methods exhibited higher errors compared to the bubbles. This discrepancy suggests that they transfer less effectively than our approach. This can be attributed to the stiffer nature of the Gelslim sensors, making them less forgiving to minor errors in force transmission compared to the bubbles.
Additionally, we observed instances of slippage between the sensor and the grasped object, which violates the assumption of sticking contact between them. Consequently, the Gelslim sensors exhibited inferior performance in tracking the desired wrenches, highlighting the challenges posed by their stiffness.

\begin{figure}[h!]
    \centering
    \includegraphics[width=0.5\textwidth]{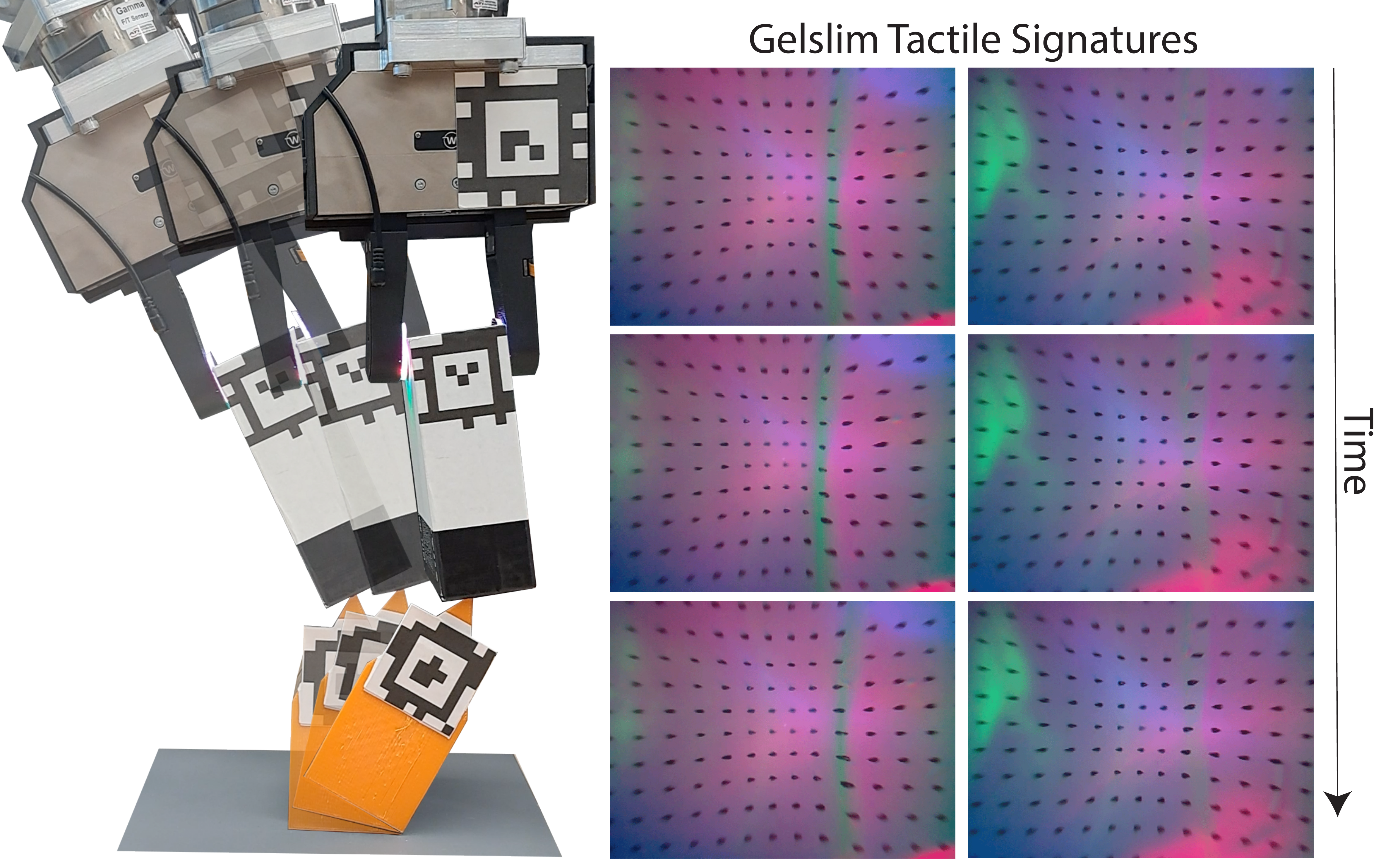}
    \caption{\textbf{Gelslim Extrinsic Pivoting Sequence:} Extrinsic pivoting sequence demonstrated using Gelslim tactile sensors, showcasing the versatility of our method which applies to both soft bubbles and more rigid sensors.}
    \label{fig:gelslim_extrinsic_pivoting}
\end{figure}

\section{Discussion, Limitations, and Future Work}

In this paper, we proposed an approach to extrinsic object manipulation leveraging tactile sensor compliance, tactile sensor measurements, and contact constraints. While our approach demonstrates successful trajectory planning for robots with compliant membranes to produce complex and dexterous behaviors, it is important to acknowledge its limitations and areas for future exploration.

Firstly, our current implementation assumes the provided object configurations are achievable and tracks them as closely as possible. However, we cannot guarantee perfect adherence to these specifications. The local planner might miss a globally optimal solution that satisfies all requirements. Addressing this limitation could involve incorporating global planning techniques within our framework.

Secondly, the optimization currently ignores the robot's dexterous workspace limitations. It assumes any configuration is attainable, which is unrealistic in practice. While we strategically place objects to maximize workspace access, joint limits might still be reached. Although incorporating dexterous workspace constraints is technically feasible, it deviates from the core focus of this work. 
Our work could be expanded to integrate workspace constraints into the trajectory optimization process, thereby restricting the robot's pose. However, as these constraints impose limitations on the robot's pose, they do not directly align with the Quadratic Program (QP) structure, where poses are typically assumed to be fixed during the QP solution. Therefore, we propose incorporating such constraints into the outer-optimization process by penalizing violations in the loss function. This can be achieved through the use of log-barrier functions, which effectively bound the robot's poses within the workspace limits. This approach is consistent with our existing strategy for preventing configurations that may lead to penetration.
Future studies could explore integrating this aspect while maintaining the core objectives.

Furthermore, our approach does not reason about the physical limitations of the bubbles in terms of achievable forces and torques. We assume continuous sticking contact between the object and the membranes. However, exceeding force or torque thresholds could lead to slippage or sensor damage. Future iterations could benefit from incorporating these physical constraints to ensure safe and reliable object manipulation.

Finally, the employed elasticity model is a simplified approximation of the actual force-deformation relationship, particularly regarding torques. While the current model yields satisfactory results, exploring higher-dimensional models with improved accuracy could further enhance performance. However, these models must be invertible to remain compatible with our optimization framework.

Building upon the foundation established in this work, several exciting avenues for future exploration present themselves:

\noindent\textit{Extending to 3D Space:} Currently, our approach operates within a planar environment. A future direction is to extend the formulation to encompass the full complexity of 3D space (SE(3)). The main elements that require modification are the friction constraints. Either the full conic friction model can be used in which case the problem becomes a Second Order Conic Program or a general Nonlinear Complementarity problem. This expansion would significantly broaden the applicability of our method to real-world manipulation tasks involving intricate object shapes and diverse robot motions.

\noindent\textit{Inferring Object Properties and Optimizing Geometry:} The optimization framework developed here possesses valuable potential beyond solely generating robot-object trajectories. By leveraging the calculated gradients, we could explore inferring unknown object properties based on observed trajectories. This capability could provide robots with valuable real-time insights into the objects they interact with, enhancing manipulation strategies. Additionally, exploring the possibility of optimizing object geometry for improved manipulability holds promise for designing objects that are specifically tailored for effective robot interaction.

\noindent\textit{Incorporating Variable Friction:} Our current assumptions treat friction coefficients as constant across the contact surfaces. A crucial improvement would be to incorporate more realistic models that capture the variability of friction properties depending on the object configuration. This would significantly enhance the accuracy and adaptability of our approach, enabling robots to handle objects with diverse and dynamic frictional characteristics.

By pursuing these future directions, we can unlock the full potential of our proposed framework, enabling robots with compliant membranes to perform even more versatile and sophisticated object manipulation tasks in complex and dynamic environments.

\section*{Acknowledgments}
We would like to thank Xili Yi, Hardik Parwana, and Mark Van der Merwe for their insightful discussions.
This research project is supported by Toyota Research Institute under the University Research Program (URP) 2.0, the NSF NRI Award 2220876, and "la Caixa" Fellowship Program (ID 100010434).

\bibliographystyle{unsrtnat}
\bibliography{references}

\begin{thebibliography}{41}
\providecommand{\natexlab}[1]{#1}
\providecommand{\url}[1]{\texttt{#1}}
\expandafter\ifx\csname urlstyle\endcsname\relax
  \providecommand{\doi}[1]{doi: #1}\else
  \providecommand{\doi}{doi: \begingroup \urlstyle{rm}\Url}\fi

\bibitem[Mason(2018)]{mason_manipulation}
Matthew~T Mason.
\newblock Toward robotic manipulation.
\newblock \emph{Annual Review of Control, Robotics, and Autonomous Systems}, 1:\penalty0 1--28, 2018.

\bibitem[Alspach et~al.(2019)Alspach, Hashimoto, Kuppuswamy, and Tedrake]{soft_bubbles}
Alex Alspach, Kunimatsu Hashimoto, Naveen Kuppuswamy, and Russ Tedrake.
\newblock Soft-bubble: A highly compliant dense geometry tactile sensor for robot manipulation.
\newblock In \emph{2019 2nd IEEE International Conference on Soft Robotics (RoboSoft)}, pages 597--604. IEEE, 2019.

\bibitem[Yuan et~al.(2017)Yuan, Dong, and Adelson]{yuan2017gelsight}
Wenzhen Yuan, Siyuan Dong, and Edward~H Adelson.
\newblock Gelsight: High-resolution robot tactile sensors for estimating geometry and force.
\newblock \emph{Sensors}, 17\penalty0 (12):\penalty0 2762, 2017.

\bibitem[Taylor et~al.(2021)Taylor, Dong, and Rodriguez]{gelslim3}
Ian Taylor, Siyuan Dong, and Alberto Rodriguez.
\newblock Gelslim3. 0: High-resolution measurement of shape, force and slip in a compact tactile-sensing finger.
\newblock \emph{arXiv preprint arXiv:2103.12269}, 2021.

\bibitem[Yamaguchi and Atkeson(2017)]{yamaguchi2017implementing}
Akihiko Yamaguchi and Christopher~G Atkeson.
\newblock Implementing tactile behaviors using fingervision.
\newblock In \emph{2017 IEEE-RAS 17th International Conference on Humanoid Robotics (Humanoids)}, pages 241--248. IEEE, 2017.

\bibitem[Lambeta et~al.(2020)Lambeta, Chou, Tian, Yang, Maloon, Most, Stroud, Santos, Byagowi, Kammerer, et~al.]{lambeta2020digit}
Mike Lambeta, Po-Wei Chou, Stephen Tian, Brian Yang, Benjamin Maloon, Victoria~Rose Most, Dave Stroud, Raymond Santos, Ahmad Byagowi, Gregg Kammerer, et~al.
\newblock Digit: A novel design for a low-cost compact high-resolution tactile sensor with application to in-hand manipulation.
\newblock \emph{IEEE Robotics and Automation Letters}, 5\penalty0 (3):\penalty0 3838--3845, 2020.

\bibitem[Rodriguez(2021)]{alberto_unstable_queen}
Alberto Rodriguez.
\newblock The unstable queen: Uncertainty, mechanics, and tactile feedback.
\newblock \emph{Science Robotics}, 6\penalty0 (54):\penalty0 eabi4667, 2021.

\bibitem[Bauza et~al.(2023)Bauza, Bronars, and Rodriguez]{bauza_tac2pose}
Maria Bauza, Antonia Bronars, and Alberto Rodriguez.
\newblock Tac2pose: Tactile object pose estimation from the first touch.
\newblock \emph{The International Journal of Robotics Research}, 42\penalty0 (13):\penalty0 1185--1209, 2023.

\bibitem[Kelestemur et~al.(2022)Kelestemur, Platt, and Padir]{tactile_pose_policy}
Tarik Kelestemur, Robert Platt, and Taskin Padir.
\newblock Tactile pose estimation and policy learning for unknown object manipulation.
\newblock \emph{arXiv preprint arXiv:2203.10685}, 2022.

\bibitem[Kuppuswamy et~al.(2019)Kuppuswamy, Castro, Phillips-Grafflin, Alspach, and Tedrake]{bubble_pose_estimation}
Naveen Kuppuswamy, Alejandro Castro, Calder Phillips-Grafflin, Alex Alspach, and Russ Tedrake.
\newblock Fast model-based contact patch and pose estimation for highly deformable dense-geometry tactile sensors.
\newblock \emph{IEEE Robotics and Automation Letters}, 5\penalty0 (2):\penalty0 1811--1818, 2019.

\bibitem[Dikhale et~al.(2022)Dikhale, Patel, Dhingra, Naramura, Hayashi, Iba, and Jamali]{visuotactile_6d_pose}
Snehal Dikhale, Karankumar Patel, Daksh Dhingra, Itoshi Naramura, Akinobu Hayashi, Soshi Iba, and Nawid Jamali.
\newblock Visuotactile 6d pose estimation of an in-hand object using vision and tactile sensor data.
\newblock \emph{IEEE Robotics and Automation Letters}, 7\penalty0 (2):\penalty0 2148--2155, 2022.

\bibitem[Chaudhury et~al.(2022)Chaudhury, Man, Yuan, and Atkeson]{vision_tactile_servoing}
Arkadeep~Narayan Chaudhury, Timothy Man, Wenzhen Yuan, and Christopher~G Atkeson.
\newblock Using collocated vision and tactile sensors for visual servoing and localization.
\newblock \emph{IEEE Robotics and Automation Letters}, 7\penalty0 (2):\penalty0 3427--3434, 2022.

\bibitem[Manuelli and Tedrake(2016)]{contact_particle_filter}
Lucas Manuelli and Russ Tedrake.
\newblock Localizing external contact using proprioceptive sensors: The contact particle filter.
\newblock In \emph{2016 IEEE/RSJ International Conference on Intelligent Robots and Systems (IROS)}, pages 5062--5069. IEEE, 2016.

\bibitem[Sipos and Fazeli(2023)]{multiscope}
Andrea Sipos and Nima Fazeli.
\newblock Multiscope: Disambiguating in-hand object poses with proprioception and tactile feedback.
\newblock \emph{arXiv preprint arXiv:2305.14204}, 2023.

\bibitem[Ma et~al.(2021)Ma, Dong, and Rodriguez]{daolin_contact_sensing}
Daolin Ma, Siyuan Dong, and Alberto Rodriguez.
\newblock Extrinsic contact sensing with relative-motion tracking from distributed tactile measurements.
\newblock In \emph{2021 IEEE international conference on robotics and automation (ICRA)}, pages 11262--11268. IEEE, 2021.

\bibitem[Kim and Rodriguez(2022)]{kim2022active}
Sangwoon Kim and Alberto Rodriguez.
\newblock Active extrinsic contact sensing: Application to general peg-in-hole insertion.
\newblock In \emph{2022 International Conference on Robotics and Automation (ICRA)}, pages 10241--10247. IEEE, 2022.

\bibitem[Higuera et~al.(2023)Higuera, Dong, Boots, and Mukadam]{neural_contact_fields}
Carolina Higuera, Siyuan Dong, Byron Boots, and Mustafa Mukadam.
\newblock Neural contact fields: Tracking extrinsic contact with tactile sensing.
\newblock In \emph{2023 IEEE International Conference on Robotics and Automation (ICRA)}, pages 12576--12582. IEEE, 2023.

\bibitem[Holladay et~al.(2021)Holladay, Lozano-P{\'e}rez, and Rodriguez]{rachel_forceful}
Rachel Holladay, Tom{\'a}s Lozano-P{\'e}rez, and Alberto Rodriguez.
\newblock Planning for multi-stage forceful manipulation.
\newblock In \emph{2021 IEEE International Conference on Robotics and Automation (ICRA)}, pages 6556--6562. IEEE, 2021.

\bibitem[Kim et~al.(2023)Kim, Jha, Romeres, Patre, and Rodriguez]{kim2023simultaneous}
Sangwoon Kim, Devesh~K Jha, Diego Romeres, Parag Patre, and Alberto Rodriguez.
\newblock Simultaneous tactile estimation and control of extrinsic contact.
\newblock \emph{arXiv preprint arXiv:2303.03385}, 2023.

\bibitem[Oller et~al.(2023)Oller, i~Lisbona, Berenson, and Fazeli]{manipulation_via_membranes}
Miquel Oller, Mireia~Planas i~Lisbona, Dmitry Berenson, and Nima Fazeli.
\newblock Manipulation via membranes: High-resolution and highly deformable tactile sensing and control.
\newblock In \emph{Conference on Robot Learning}, pages 1850--1859. PMLR, 2023.

\bibitem[Masterjohn et~al.(2022)Masterjohn, Guoy, Shepherd, and Castro]{bubble_hydroelastic_model}
Joseph Masterjohn, Damrong Guoy, John Shepherd, and Alejandro Castro.
\newblock Velocity level approximation of pressure field contact patches.
\newblock \emph{IEEE Robotics and Automation Letters}, 7\penalty0 (4):\penalty0 11593--11600, 2022.

\bibitem[Suh et~al.(2022)Suh, Kuppuswamy, Pang, Mitiguy, Alspach, and Tedrake]{seed}
HJ~Terry Suh, Naveen Kuppuswamy, Tao Pang, Paul Mitiguy, Alex Alspach, and Russ Tedrake.
\newblock Seed: Series elastic end effectors in 6d for visuotactile tool use.
\newblock In \emph{2022 IEEE/RSJ International Conference on Intelligent Robots and Systems (IROS)}, pages 4684--4691. IEEE, 2022.

\bibitem[Posa et~al.(2014)Posa, Cantu, and Tedrake]{posa2014direct}
Michael Posa, Cecilia Cantu, and Russ Tedrake.
\newblock A direct method for trajectory optimization of rigid bodies through contact.
\newblock \emph{The International Journal of Robotics Research}, 33\penalty0 (1):\penalty0 69--81, 2014.

\bibitem[Manchester and Kuindersma(2020)]{manchester2020variational}
Zachary Manchester and Scott Kuindersma.
\newblock Variational contact-implicit trajectory optimization.
\newblock In \emph{Robotics Research: The 18th International Symposium ISRR}, pages 985--1000. Springer, 2020.

\bibitem[Mordatch et~al.(2012)Mordatch, Popovi{\'c}, and Todorov]{mordatch2012contact}
Igor Mordatch, Zoran Popovi{\'c}, and Emanuel Todorov.
\newblock Contact-invariant optimization for hand manipulation.
\newblock In \emph{Proceedings of the ACM SIGGRAPH/Eurographics symposium on computer animation}, pages 137--144, 2012.

\bibitem[Aceituno-Cabezas and Rodriguez(2020)]{bernardo_quasidynamic_contact}
Bernardo Aceituno-Cabezas and Alberto Rodriguez.
\newblock A global quasi-dynamic model for contact-trajectory optimization in manipulation.
\newblock 2020.

\bibitem[Chavan-Dafle et~al.(2020)Chavan-Dafle, Holladay, and Rodriguez]{nikhil_planar_motion_cones}
Nikhil Chavan-Dafle, Rachel Holladay, and Alberto Rodriguez.
\newblock Planar in-hand manipulation via motion cones.
\newblock \emph{The International Journal of Robotics Research}, 39\penalty0 (2-3):\penalty0 163--182, 2020.

\bibitem[Hou et~al.(2018)Hou, Jia, and Mason]{fast_pivoting_planning}
Yifan Hou, Zhenzhong Jia, and Matthew~T Mason.
\newblock Fast planning for 3d any-pose-reorienting using pivoting.
\newblock In \emph{2018 IEEE International Conference on Robotics and Automation (ICRA)}, pages 1631--1638. IEEE, 2018.

\bibitem[Hogan et~al.(2020)Hogan, Ballester, Dong, and Rodriguez]{hogan_tactile_dexterity}
Francois~R Hogan, Jose Ballester, Siyuan Dong, and Alberto Rodriguez.
\newblock Tactile dexterity: Manipulation primitives with tactile feedback.
\newblock In \emph{2020 IEEE international conference on robotics and automation (ICRA)}, pages 8863--8869. IEEE, 2020.

\bibitem[Cheng et~al.(2023)Cheng, Patil, Temel, Kroemer, and Mason]{mason_hierarchical_contact}
Xianyi Cheng, Sarvesh Patil, Zeynep Temel, Oliver Kroemer, and Matthew~T Mason.
\newblock Enhancing dexterity in robotic manipulation via hierarchical contact exploration.
\newblock \emph{IEEE Robotics and Automation Letters}, 9\penalty0 (1):\penalty0 390--397, 2023.

\bibitem[Dafle et~al.(2014)Dafle, Rodriguez, Paolini, Tang, Srinivasa, Erdmann, Mason, Lundberg, Staab, and Fuhlbrigge]{nikhil_extrinsic_dexterity}
Nikhil~Chavan Dafle, Alberto Rodriguez, Robert Paolini, Bowei Tang, Siddhartha~S Srinivasa, Michael Erdmann, Matthew~T Mason, Ivan Lundberg, Harald Staab, and Thomas Fuhlbrigge.
\newblock Extrinsic dexterity: In-hand manipulation with external forces.
\newblock In \emph{2014 IEEE International Conference on Robotics and Automation (ICRA)}, pages 1578--1585. IEEE, 2014.

\bibitem[Chavan-Dafle and Rodriguez(2018)]{nikhil_stable_prehensile_pushing}
Nikhil Chavan-Dafle and Alberto Rodriguez.
\newblock Stable prehensile pushing: In-hand manipulation with alternating sticking contacts.
\newblock In \emph{2018 IEEE International Conference on Robotics and Automation (ICRA)}, pages 254--261. IEEE, 2018.

\bibitem[Hou et~al.(2020)Hou, Jia, and Mason]{manipulation_shared_grasping}
Yifan Hou, Zhenzhong Jia, and Matthew~T Mason.
\newblock Manipulation with shared grasping.
\newblock \emph{arXiv preprint arXiv:2006.02996}, 2020.

\bibitem[Shirai et~al.(2022)Shirai, Jha, Raghunathan, and Romeres]{robust_pivoting_shirai}
Yuki Shirai, Devesh~K Jha, Arvind~U Raghunathan, and Diego Romeres.
\newblock Robust pivoting: Exploiting frictional stability using bilevel optimization.
\newblock In \emph{2022 International Conference on Robotics and Automation (ICRA)}, pages 992--998. IEEE, 2022.

\bibitem[Doshi et~al.(2022)Doshi, Taylor, and Rodriguez]{doshi2022manipulation}
Neel Doshi, Orion Taylor, and Alberto Rodriguez.
\newblock Manipulation of unknown objects via contact configuration regulation.
\newblock In \emph{2022 International Conference on Robotics and Automation (ICRA)}, pages 2693--2699. IEEE, 2022.

\bibitem[Taylor et~al.(2023)Taylor, Doshi, and Rodriguez]{taylor2023object}
Orion Taylor, Neel Doshi, and Alberto Rodriguez.
\newblock Object manipulation through contact configuration regulation: multiple and intermittent contacts.
\newblock In \emph{2023 IEEE/RSJ International Conference on Intelligent Robots and Systems (IROS)}, pages 8735--8743. IEEE, 2023.

\bibitem[Shirai et~al.(2023)Shirai, Jha, Raghunathan, and Hong]{tactile_tool_manipulation}
Yuki Shirai, Devesh~K Jha, Arvind~U Raghunathan, and Dennis Hong.
\newblock Tactile tool manipulation.
\newblock \emph{arXiv preprint arXiv:2301.06698}, 2023.

\bibitem[Amos and Kolter(2017)]{optnet}
Brandon Amos and J~Zico Kolter.
\newblock Optnet: Differentiable optimization as a layer in neural networks.
\newblock In \emph{International Conference on Machine Learning}, pages 136--145. PMLR, 2017.

\bibitem[Agrawal et~al.(2019)Agrawal, Amos, Barratt, Boyd, Diamond, and Kolter]{cvxpylayers}
Akshay Agrawal, Brandon Amos, Shane Barratt, Stephen Boyd, Steven Diamond, and J~Zico Kolter.
\newblock Differentiable convex optimization layers.
\newblock \emph{Advances in neural information processing systems}, 32, 2019.

\bibitem[Williams et~al.(2017)Williams, Wagener, Goldfain, Drews, Rehg, Boots, and Theodorou]{mppi}
Grady Williams, Nolan Wagener, Brian Goldfain, Paul Drews, James~M Rehg, Byron Boots, and Evangelos~A Theodorou.
\newblock Information theoretic mpc for model-based reinforcement learning.
\newblock In \emph{2017 IEEE International Conference on Robotics and Automation (ICRA)}, pages 1714--1721. IEEE, 2017.

\bibitem[Pinneri et~al.(2020)Pinneri, Sawant, Blaes, Achterhold, Stueckler, Rolinek, and Martius]{icem}
Cristina Pinneri, Shambhuraj Sawant, Sebastian Blaes, Jan Achterhold, Joerg Stueckler, Michal Rolinek, and Georg Martius.
\newblock Sample-efficient cross-entropy method for real-time planning.
\newblock In \emph{Conference on Robot Learning 2020}, 2020.

\end{thebibliography}

\end{document}